\documentclass[11pt]{article}

\usepackage[final]{acl}
\usepackage{times}
\usepackage{latexsym}
\usepackage[T1]{fontenc}
\usepackage[utf8]{inputenc}
\usepackage{microtype}
\usepackage{inconsolata}
\usepackage{graphicx}
\usepackage{booktabs}
\usepackage{array}
\usepackage{amsmath}
\usepackage{enumitem}
\usepackage{fvextra}
\usepackage{placeins}

\hypersetup{
  pdftitle={Compile by Training: Turning Natural-Language Specifications into Local Neural Functions},
  pdfauthor={Yuntian Deng, Pengyu Nie, and Stuart Shieber}
}

\title{Compile by Training: Turning Natural-Language\\ Specifications into Local Neural Functions}

\author{
  Yuntian Deng \\
  University of Waterloo \\
  \texttt{yuntian@uwaterloo.ca}
  \And
  Pengyu Nie \\
  University of Waterloo \\
  \texttt{pynie@uwaterloo.ca}
  \And
  Stuart Shieber \\
  Harvard University \\
  \texttt{shieber@seas.harvard.edu}
}

\begin{document}
\maketitle

\begin{abstract}
Many recurring text functions are easy to describe but difficult to implement with rules, while calling a large remote model for every input introduces repeated cost, latency, and dependency on a provider. We present \emph{compile by training}, which turns a natural-language specification into a reusable neural function. At compile time, teacher models generate task-specific examples that are used to train a small adapter for a compact interpreter. The resulting function runs without the teachers and can be stored, versioned, and composed like ordinary software. On FuzzyBench-Hard, a subset on which the Program-as-Weights fast compiler produced no exact matches, compile by training reaches 83.6\% semantic accuracy. This higher accuracy comes with a higher compile-time cost: roughly a minute rather than seconds for the fast compiler. We deploy the compiler in a public interactive service and demonstrate compiled functions in a multi-site website helper, a language-controlled 3D avatar, and a bidirectional English--Claudish translator.
\end{abstract}

\section{Introduction}
Consider an email-triage function that maps ``Signature needed by EOD'' to \texttt{immediate}, but maps a newsletter to \texttt{wait}. The intended behavior is easy to describe and useful across thousands of messages, yet tedious to encode as rules. A general-purpose remote LLM can perform the task, but calling it for every message repeatedly incurs network latency, provider cost, and dependence on an external service. Many recurring text functions fall between these two options: they are too fuzzy for conventional code, but too narrow and frequent to justify a large-model call at every invocation.

We propose \emph{compile by training}: use large models once to build a smaller, reusable neural function. A developer writes a natural-language specification, teacher models generate examples of the desired behavior, and gradient descent uses those examples to specialize a compact interpreter. Compilation takes roughly a minute, but the resulting function can then process new inputs without calling the teachers. In this view, adaptation becomes a software build step, and large language models act as tool builders rather than run-time
dependencies.

Our system builds on Program-as-Weights (PAW) \citep{zhang2026programs}, in
which a neural program specializes a shared local interpreter. PAW introduced an amortized
compiler that predicts such a program in a single forward pass. This makes
compilation fast, but spends the same fixed amount of computation on every
function. Compile by training retains the same program format and run-time
interface, uses the amortized prediction as a starting point, and then invests
additional computation in teacher synthesis and specification-specific
optimization. The resulting programs can still be versioned, cached, and composed with ordinary software.

This approach raises three practical questions. First, does the additional
compile-time investment produce better functions than one-shot
weight prediction? Second, can a minute-scale build remain usable in an
interactive service? Third, can independently compiled neural functions serve
as components in larger applications? We evaluate correctness on
FuzzyBench-Hard, a subset of examples for which the PAW fast compiler produced
no exact matches. We also measure latency and responsiveness in our deployed
compiler and demonstrate composition in a multi-site website helper, a
language-controlled 3D avatar, and a bidirectional English--Claudish translator.
An online demo is available at
\url{https://programasweights.com/playground?compiler=paw-ft-bs48}.

\begin{figure*}[!t]
  \centering
  \includegraphics[width=0.9\textwidth]{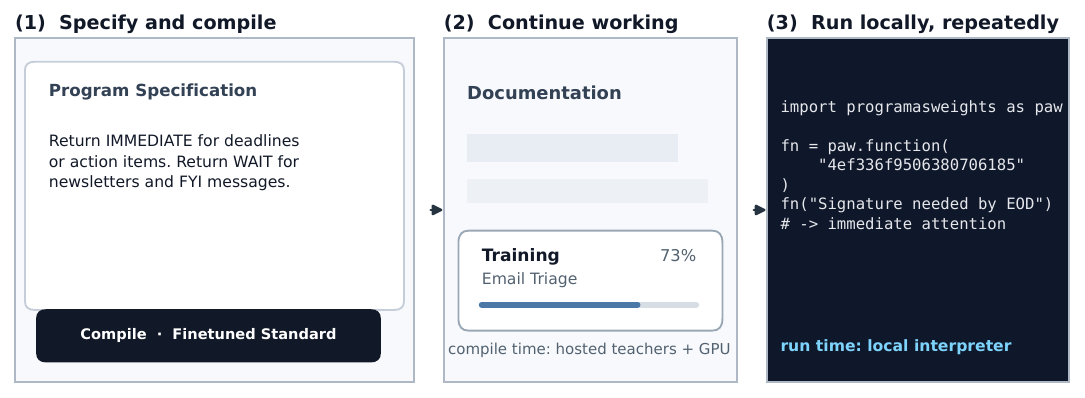}
  \caption{\textbf{One specification, one compile job, one local function.}
  The running email-triage example is entered in the public playground, compiled
  asynchronously while the user continues browsing, and then invoked through the
  local SDK. The teacher and GPU are compile-time dependencies; repeated function
  calls use the packaged program.}
  \label{fig:workflow}
\end{figure*}

\section{System Overview}

The system separates building a neural function from using it. A developer first describes the desired behavior in natural language and compiles that specification into a reusable program. The completed program can then be invoked repeatedly on new inputs without calling the teacher models again.

Formally, the system exposes the interface
\[
p_s = \texttt{Compile}(s), \qquad
\hat{y} = \texttt{Run}(p_s, x),
\]
where $s$ is a natural-language function specification, $p_s$ is the compiled program, $x$ is a new input, and $\hat{y}$ is the program's output. A shared frozen language model serves as the \emph{interpreter}, while each compiled program supplies the adapter and prompt that specialize it for one function.

Figure~\ref{fig:workflow} shows the corresponding user workflow:
\begin{enumerate}[leftmargin=*,nosep]
\item \textbf{Specify.} Describe a text-to-text function in natural language.
\item \textbf{Compile.} Submit a build job and monitor its queue and training progress.
\item \textbf{Run.} Open or download the completed program and invoke it on new inputs through the SDK.
\end{enumerate}

The downloaded \texttt{.paw} artifact packages the specification and the components needed to specialize the shared interpreter. It can be stored, versioned, cached, and reused like an ordinary software artifact. During execution, the SDK combines each new input with the program's run-time prompt and adapter, then runs the shared interpreter.

Compilation is a hosted process: the function specification is sent to the PAW service and teacher APIs to generate supervision and train the program. By contrast, local SDK execution does not send future inputs to PAW or the teachers.

\section{Compile by Training}

Compile by training turns a natural-language specification into a reusable neural program in two stages. First, teacher models synthesize examples of the desired function. Second, those examples specialize a lightweight adapter for a shared interpreter, which is then packaged for future execution.

\subsection{From Specification to Supervision}

A specification describes the desired behavior but does not provide enough labeled examples to train it directly. We therefore use one or more teacher models to synthesize a task-specific dataset
\[
D_s = \{(x_i,y_i)\}_{i=1}^{n} \sim T(s),
\]
where $s$ is the specification and each pair illustrates how the function should map an input $x_i$ to an output $y_i$.

Teacher requests use a structured JSON format so that generated examples can enter the training pipeline automatically. The compiler validates each response and rejects malformed or incomplete batches before making accepted examples available to the trainer. Our public service combines a lower-cost teacher, which supplies most of the examples, with a larger teacher that contributes complementary supervision.

For example, for a specification that extracts arXiv identifiers from \texttt{/pdf/} links while ignoring \texttt{/abs/} links, a teacher-generated example is:
\begin{quote}\small
\textbf{Input:} Check these papers: \url{https://arxiv.org/pdf/2607.02512.pdf} and \url{https://arxiv.org/abs/2507.08800}.\\
\textbf{Output:} \texttt{["2607.02512"]}
\end{quote}

\subsection{Specialization and Packaging}

Training a separate full model for every specification would make compilation and storage prohibitively expensive. Instead, all programs share a frozen Qwen3-0.6B interpreter~\citep{qwen3}, and each function is represented by a lightweight LoRA adapter~\citep{hu2021lora} and a run-time scaffold. Here, the scaffold is a compiler-generated prompt template that encodes the user's free-form specification as structured task instructions and examples and leaves a placeholder for the run-time input.

The amortized PAW compiler provides the initial adapter parameters
$\theta_s^{(0)}$ and scaffold $r_s$. This prediction takes seconds and provides a function-specific starting point, which is then refined using the synthesized dataset. Training minimizes
\[
\mathcal{L}(\theta_s)
= \sum_{(x,y)\in D_s}
-\log p_{\theta_s}\!\left(y \mid r_s(x)\right).
\]

\begin{figure*}[!t]
  \centering
  \includegraphics[width=0.9\textwidth]{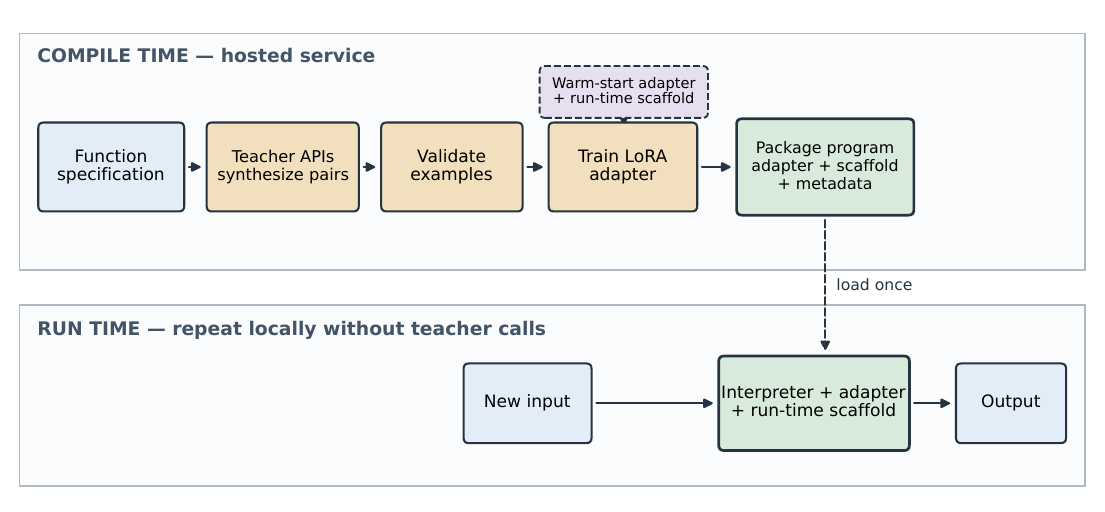}
  \caption{\textbf{Compile by training.} At compile time, teacher models turn a
  specification into validated supervision, which specializes a LoRA adapter
  for a compact frozen interpreter. The compiler packages the adapter and
  run-time scaffold into a reusable program, which handles new inputs at run time without teacher calls.}
  \label{fig:method}
\end{figure*}

After optimization, the compiler packages the adapter $\theta_s$, scaffold $r_s$, original specification, and interpreter metadata into the program $p_s$. Figure~\ref{fig:method} summarizes this path from a natural-language description to a reusable artifact.

\paragraph{Public configuration.}
The public \emph{Finetuned Standard} compiler uses a quantized Qwen3-0.6B interpreter, mixed teachers, a rank-64 LoRA adapter with alpha 16, an amortized-compiler warm start, and a 100-step cosine schedule. Appendix Table~\ref{tab:compact-recipe} gives the complete recipe.

\section{Interactive Minute-Scale Compilation}

Compile by training takes tens of seconds rather than a single model call. To make this build step usable interactively, the system must both shorten the critical path and let users continue working while compilation proceeds. We address these requirements at three levels: overlapping synthesis with training, coordinating work across shared workers, and presenting each compile as a persistent background job.

\subsection{Overlapping Synthesis and Training}

In a sequential pipeline, the compiler would wait for all teacher-generated examples before loading the interpreter and beginning optimization. This leaves the GPU idle while waiting for the slowest teacher responses. Our deployed \emph{streaming compile path} instead starts teacher requests, model loading, and training concurrently.

Training begins as soon as the examples required for its first batch are available and blocks only if it catches up with synthesis. Because teacher responses may arrive at different times, the scheduler assigns incoming examples to open slots needed by earlier batches first, without changing the accepted examples or their prescribed training order. This overlap shortens the critical path because teacher synthesis is substantially slower and more variable than an individual training step.

\subsection{Coordinating Jobs and Reusing Work}

\begin{figure*}[t]
  \centering
  \includegraphics[width=0.9\textwidth]{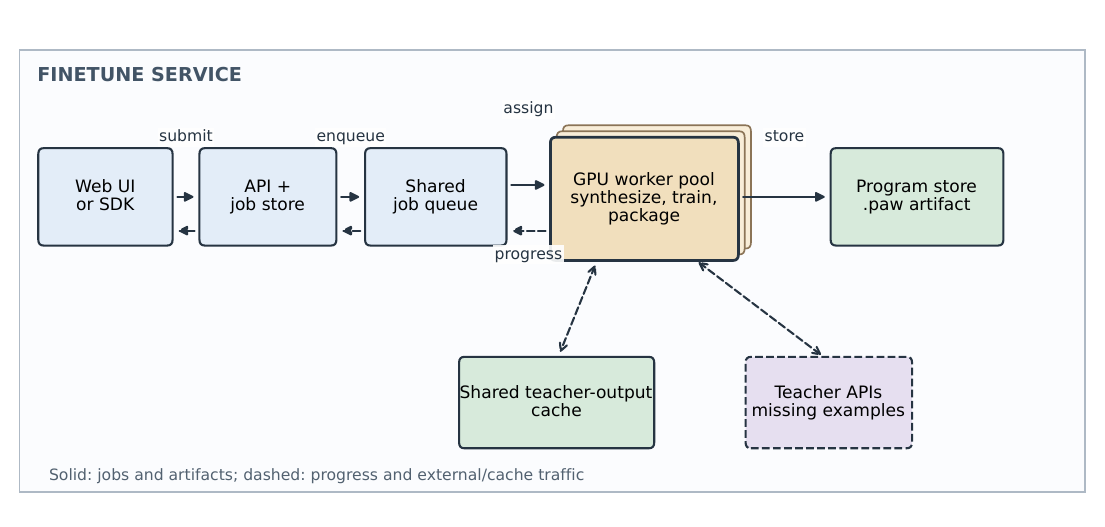}
  \caption{\textbf{Deployed finetune service.} The API submits a job to a shared
  queue, which assigns it to an available GPU. Workers reuse cached teacher
  outputs, call teachers for missing examples, execute the
  compiler in Figure~\ref{fig:method}, and report progress. Completed artifacts
  are stored centrally and returned through the same job record. Solid arrows
  are requests or artifacts; dashed arrows are status and cache traffic.}
  \label{fig:service}
\end{figure*}

The service separates job coordination from the compilation workers that perform synthesis and training. The API maintains a persistent record of each submitted job, while a shared queue dispatches pending jobs to available GPU workers. Before requesting new examples, workers check the cache and reuse matching teacher outputs. Completed programs are written to a shared artifact store and linked back to the persistent job record. Figure~\ref{fig:service} summarizes this service architecture.

This separation allows the API to remain responsive while workers execute long-running builds, and it keeps available GPUs supplied with compilation jobs.

\subsection{Interaction Design}

The interface treats compilation as a background software build rather than a blocking form submission. After submitting a specification, users can continue browsing while the interface reports the job's queue position and training progress.

The job and its progress persist across page navigation and reloads, as illustrated in Figure~\ref{fig:workflow}(2). When compilation finishes, the same job record provides access to the completed program.

\section{Evaluation and Validation}
\label{sec:evaluation}

\subsection{Metric and benchmark}

For many tasks, multiple outputs can all be valid. For example, two JSON outputs
may contain the same values while differing only in whitespace or key order,
yet an exact-match metric would treat them as different. We therefore report
\emph{LLM Exact Match} (LEM): the fraction of predictions that an LLM judge
considers semantically correct according to the specification. Each judgment is based on
the specification, input, reference output, and model prediction. The selected
GPT-5.5 judge reaches 0.977 accuracy and Cohen's
\(\kappa=0.946\) against 128 author labels
(Appendix Table~\ref{tab:compact-grader}). Appendix~\ref{sec:compact-lem-prompt}
gives the full grader prompt and user template.

We evaluate on \emph{FuzzyBench-Hard}, a subset of the FuzzyBench benchmark used to evaluate PAW~\citep{zhang2026programs}. The subset consists of specifications for which the PAW fast compiler produced no exact matches. As shown below, however, the fast compiler has a nonzero LEM score because some predictions are semantically correct even though they do not exactly match the reference outputs.

\subsection{Does training improve correctness?}
\label{sec:correctness}

\begin{figure}[t]
  \centering
  \includegraphics[width=\columnwidth]{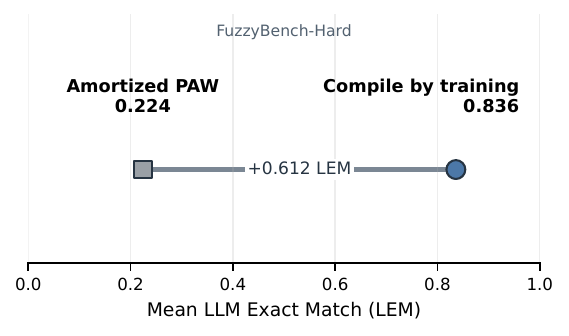}
  \caption{\textbf{Compile-time optimization buys a large correctness gain.}
  Mean LEM on FuzzyBench-Hard for PAW's fast amortized compiler and our
  compile-by-training compiler.}
  \label{fig:focus}
\end{figure}

Figure~\ref{fig:focus} compares the two compilers on FuzzyBench-Hard.
Compile by training improves mean LEM by 0.612
absolute, from 0.224 to 0.836. This improvement comes at the cost of longer
compilation: 50.9 seconds, compared with 3.5 seconds for PAW's fast amortized
compiler.

\subsection{How do supervision choices affect correctness?}

Table~\ref{tab:supervision-sweeps} reports existing sweeps on the development
specifications. With 3600 unique pairs repeated
to form a 6400-example training set, a 2:1 mixture of GPT-5.4-mini and GPT-5.5 supervision raises
mean LEM from 0.746 to 0.851 relative to GPT-5.4-mini alone. In a separate
data-scaling sweep, mean LEM rises from 0.821 with 1440 unique pairs to 0.836
with 2400, remains 0.836 with 3600, and reaches 0.866 with 7200. Appendix
\ref{sec:supervision-protocol} provides details of the sweep.

\begin{center}
\begin{minipage}{0.98\columnwidth}
  \centering
  \footnotesize
  \begin{tabular}{@{}lp{0.55\columnwidth}r@{}}
    \toprule
    sweep & setting & LEM \\
    \midrule
    teacher mix & mini only (3600/0) & 0.746 \\
    teacher mix & 2:1 mini/GPT-5.5 (2400/1200) & 0.851 \\
    data scaling & 1440 unique pairs & 0.821 \\
    data scaling & 2400 unique pairs & 0.836 \\
    data scaling & 3600 unique pairs & 0.836 \\
    data scaling & 7200 unique pairs & 0.866 \\
    \bottomrule
  \end{tabular}
  \captionof{table}{Mean LEM across controlled teacher-mixture and data-scaling sweeps.}
  \label{tab:supervision-sweeps}
\end{minipage}
\end{center}

\begin{figure*}[!t]
  \centering
  \includegraphics[width=0.94\textwidth]{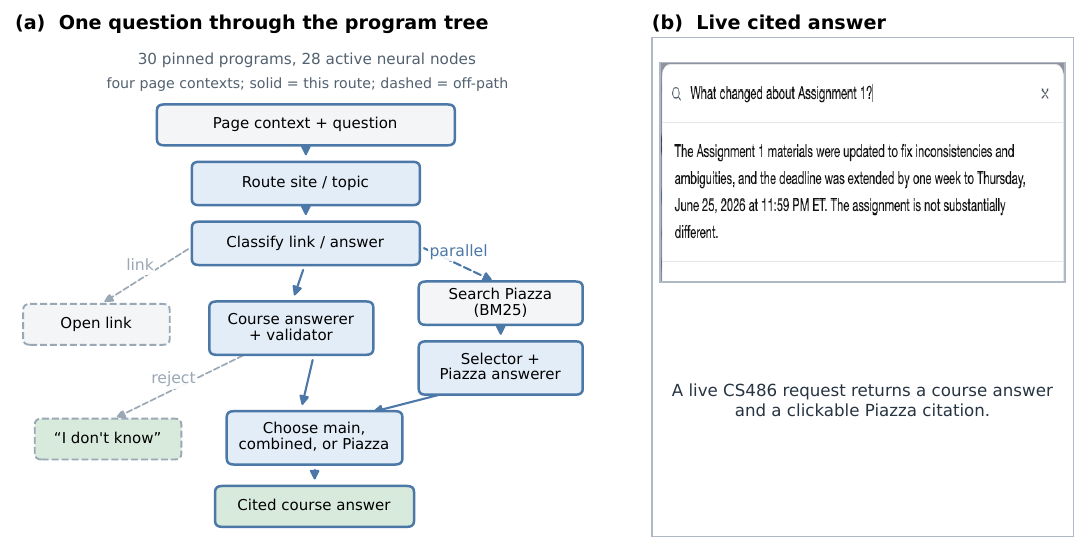}
  \caption{\textbf{A website helper as a program tree.}
  \textbf{(a)} A page-aware question is routed through finetuned classifiers,
  answerers, selectors, and validators alongside deterministic links and
  retrieval. The highlighted CS486 path combines the course answer with a Piazza search and merge.
  \textbf{(b)} The live course helper returns the result.}
  \label{fig:paw-helper}
\end{figure*}

\subsection{Is compilation fast enough for interaction?}

We measured compile latency at service launch in May 2026. A cold compile
(with no cached teacher outputs) of the same representative specification took
50.9\,s on a B300, 68.2\,s on an H200, and 99.2\,s on an RTX GPU. In a load test in which four compile
jobs were submitted concurrently, all four completed with a mean queue wait of
1.01\,s and even utilization across workers. Because teacher synthesis
dominates end-to-end latency, overlapping synthesis with training directly
shortens the critical path of our service.

\section{Deployment and Applications}
\label{sec:applications}

\subsection{Composing many compiled functions}

FuzzyBench-Hard evaluates compiled functions individually. Real applications
may need several compiled functions to work together with ordinary code.
Paw-helper demonstrates this setting in a deployed website assistant.

Paw-helper\footnote{\url{https://github.com/programasweights/paw-helper}}
separates a generic pipeline executor from a site-specific content pack. The
deployed pack contains 30 compiled programs; 28 participate in live
routing, while two support evaluation and backward compatibility. One backend
serves four websites:
\href{https://yuntiandeng.com}{the first author's personal website},
\href{https://yuntiandeng.com/teaching/spring2026/cs486-introduction-to-artificial-intelligence/\#ask}{the site for his introductory AI course at the University of Waterloo (CS 486)},
\href{https://neural-os.com}{NeuralOS}~\citep{rivard2026neuralos}, and
\href{https://programasweights.com}{the PAW website}. The router receives the
current website and page as context and sends each question through only a
small part of the tree.

When a student asks, ``What changed about Assignment 1?'', paw-helper first
decides whether the question calls for a link or a written answer. Since this
one calls for a written answer, one compiled function drafts a response using
the course page. In parallel, ordinary BM25 retrieval searches Piazza, another
compiled function answers using the most relevant post, and a final function
decides whether to return the course answer, the Piazza answer, or both.
Figure~\ref{fig:paw-helper} shows this process and the resulting answer.

This illustrates the general division of labor: compiled functions make fuzzy
decisions, while ordinary code handles exact operations such as retrieval,
caching, and branch control.

\begin{figure*}[!t]
  \centering
  \includegraphics[width=0.9\textwidth]{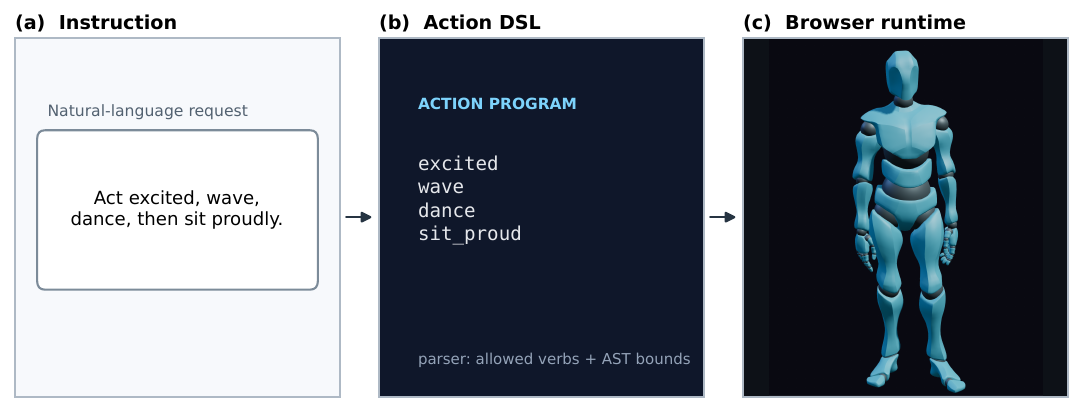}
  \caption{\textbf{From language to executable behavior.} A finetuned PAW
  program maps a natural-language command to an action program; deterministic
  browser code parses and validates the DSL before rendering the resulting
  motion. In the live web demo, DSL interpretation and 3D rendering run in the
  browser.}
  \label{fig:avatar}
\end{figure*}

\subsection{Generating executable behavior}
\label{sec:applications:avatar}

To demonstrate an output richer than a label or free-form answer, Avatar
Director\footnote{\url{https://programasweights.com/avatar}} lets a user control
a 3D character with natural-language instructions. For example, a user can ask
it to ``jump twice, then dance,'' and the avatar performs those actions in
order. A finetuned PAW program translates the natural-language instruction into
a small action DSL. The DSL supports sequences, durations, repetition, and
compatible parallel motions, and the browser validates and executes it to
animate the character. Users can therefore describe the motion in their own
words rather than writing the action program themselves. On 44 hand-authored
validation instructions, the program produced the expected action structure in
43 cases.

Figure~\ref{fig:avatar} shows the generated action program and the resulting
browser animation.

\subsection{Translating between English and Claudish}

\emph{Claudish} has recently emerged as an informal name for the
distinctive prose style associated with Claude Code.\footnote{Our demo was
inspired by the earlier
\href{https://github.com/gvzdv/claudish-to-english}{Claudish-to-English
project}; see also
\href{https://www.usermag.co/p/vibe-coders-now-have-their-own-language}
{Lorenz (2026)} for a broader discussion of the term.}
The style often uses explicit contrasts; metaphors of \emph{gates},
\emph{boundaries}, and \emph{load-bearing} structure; and hyphenated
constructions such as \emph{X-shaped} and \emph{X-gated}. We built
a live bidirectional translation service between Claudish and plain
English.\footnote{\url{https://programasweights.com/claudish}}

To build the translator, we wrote one natural-language specification for
each direction. The English-to-Claudish specification asks the program to
preserve the input's meaning while adopting the style's characteristic
vocabulary, compounds, rhetorical patterns, and sentence structures. For the
reverse Claudish-to-English direction, the specification asks the program to
rewrite the input in direct English, removing redundant contrasts and structural
metaphors while preserving its meaning. Compile by training synthesized examples
from these specifications and finetuned one adapter per direction for the
shared 0.6B interpreter. The resulting two programs power the live service.

\begin{figure*}[!t]
  \centering
  \includegraphics[width=0.8\textwidth]{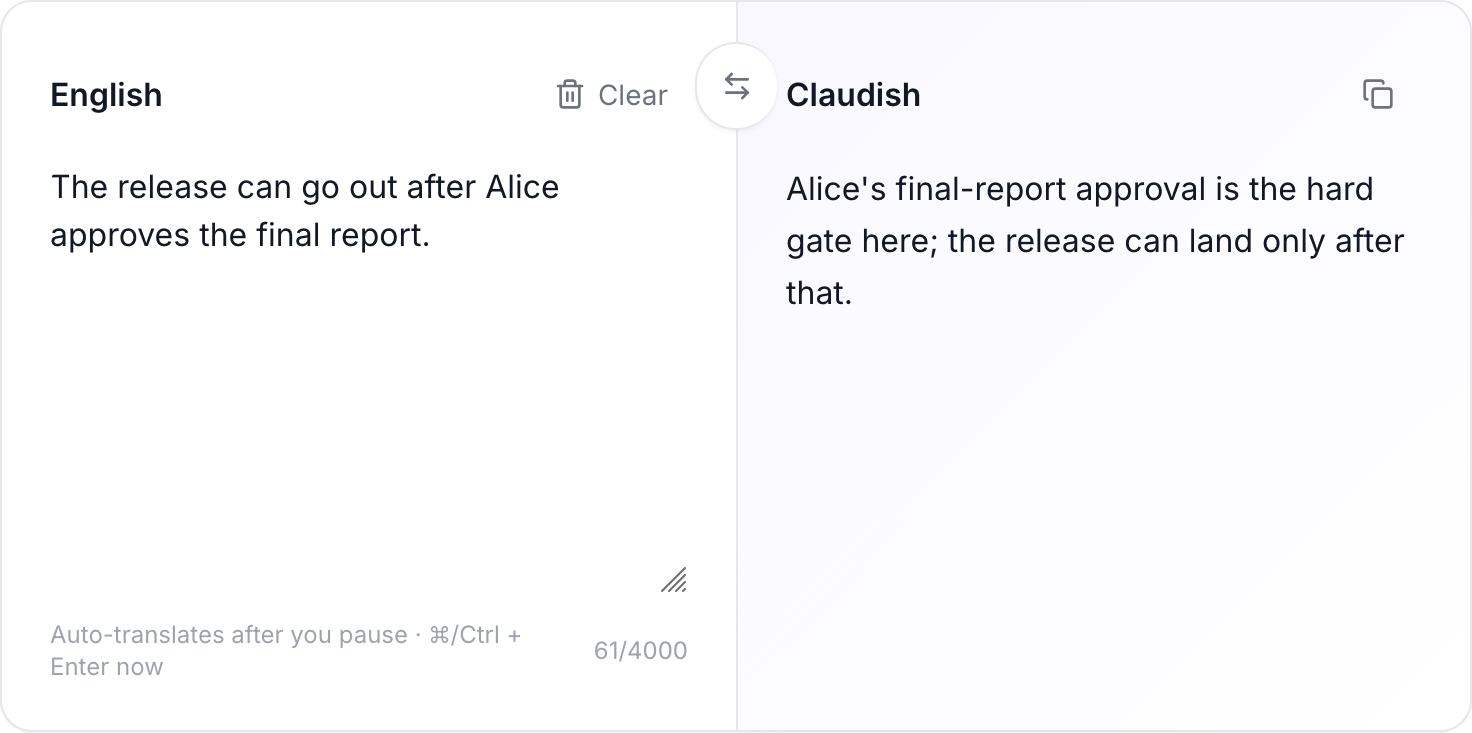}
  \caption{\textbf{English to Claudish.} The live interface applies the
  English-to-Claudish PAW program to a plain-English sentence. Switching
  direction invokes the separately finetuned Claudish-to-English program.}
  \label{fig:claudish}
\end{figure*}

Figure~\ref{fig:claudish} shows one translation in the web interface. Both
programs can also be downloaded and run locally; their specifications and
supporting code are available in a public
repository.\footnote{\url{https://github.com/programasweights/claudish}}
Between its public launch on August 22 and September 2, 2026, the web demo
completed 100,747 successful translation requests.

\section{Related Work}

\paragraph{Synthetic supervision and parameter-efficient adaptation.}
Large models can generate instruction-following examples for training smaller
models, while knowledge distillation trains a smaller model to imitate a larger
teacher~\citep{wang2023selfinstruct,taori2023alpaca,hinton2015distilling,hsieh2023distilling}.
Compile by training applies both ideas to one user-defined function: teachers
generate input-output pairs from its specification, and those pairs train a LoRA
adapter for the compact interpreter. LoRA~\citep{hu2021lora},
adapters~\citep{houlsby2019parameter}, prefix/prompt
tuning~\citep{li2021prefix,lester2021power}, QLoRA~\citep{dettmers2023qlora},
and AdaLoRA~\citep{zhang2023adalora} reduce adaptation cost by updating only a
small set of parameters.
Prompt2Model~\citep{viswanathan2023prompt2model} turns a task description into a finetune pipeline, and LoRA Land~\citep{zhao2024loraland} serves many task-specific LoRA adapters on a shared base model. Our system produces a versioned program for a shared frozen interpreter and uses an amortized compiler to initialize specification-specific training.

\section{Conclusion}

We introduce compile by training, which compiles a neural program from a
user-provided function description. It first uses large language models to
synthesize examples from the description, then finetunes a LoRA adapter to
specialize a smaller interpreter model. This adds another point to PAW's
speed--accuracy tradeoff: the fast compiler generates a LoRA adapter in a single
forward pass and takes seconds, while compile by training spends roughly a
minute to obtain higher accuracy. On FuzzyBench-Hard, a set of tasks for which
the PAW fast compiler produced no exact matches, compile by training reaches
83.6\% semantic accuracy. We added it to the public PAW demo as another option
in the same compiler interface and used the resulting programs to build a
multi-site website helper, a language-controlled 3D avatar, and a bidirectional
English--Claudish translator.

\section{Limitations}

Synthetic supervision may inherit teacher errors. Applications that require
guaranteed correctness should validate outputs or retain deterministic control
paths.
Our application evidence focuses on composition and structured execution;
systematic user studies are future work.

\section*{Acknowledgments}
Yuntian Deng acknowledges support from NSERC under funding reference RGPIN-2024-05178 and from a Google Research Award for Machine Learning Research and Education with TPUs.
Pengyu Nie acknowledges support from NSERC under funding reference RGPIN-2024-04909.

\bibliography{custom}

\clearpage
\appendix
\section{Additional Experimental Details}\label{sec:appendix-details}

This appendix collects the details most useful for interpreting the main paper.

\begin{center}
\begin{minipage}{0.98\columnwidth}
\centering
\footnotesize
\begin{tabular}{@{}p{0.24\columnwidth}p{0.68\columnwidth}@{}}
\toprule
item & value \\
\midrule
interpreter & Qwen3-0.6B with a quantized local runtime \\
teachers & GPT-5.4-mini and GPT-5.5 in a 2:1 mixture \\
data & 2400 unique pairs expanded to 6400 training examples \\
optimization & batch 48; 100 steps; cosine learning-rate decay \\
adapter & LoRA rank 64, alpha 16; hypernetwork initialization \\
execution & shuffled batches with synthesis and training overlapped \\
\bottomrule
\end{tabular}
\captionof{table}{Public Finetuned Standard configuration.}
\label{tab:compact-recipe}
\end{minipage}
\end{center}

\section{LEM Grader Summary}

LLM Exact Match (LEM) is the mean of binary, specification-grounded semantic-correctness judgments over (specification, input, reference output, model output). The selected GPT-5.5 grader tolerates formatting differences but rejects errors in values, order, structure, or logic, as well as inappropriate refusals.

\begin{center}
\begin{minipage}{0.98\columnwidth}
\centering
\scriptsize
\begin{tabular}{@{}lccccc@{}}
\toprule
grader & n & acc & FPR & FNR & kappa \\
\midrule
GPT-5.5 & 128 & 0.977 & 0.025 & 0.023 & 0.946 \\
GPT-5.4-mini & 128 & 0.938 & 0.05 & 0.068 & 0.858 \\
\bottomrule
\end{tabular}
\captionof{table}{Compact LEM grader validation.}
\label{tab:compact-grader}
\end{minipage}
\end{center}

\section{Supervision Sweep Protocol}
\label{sec:supervision-protocol}

Main Table~\ref{tab:supervision-sweeps} reports existing sweeps on the development specifications. The teacher-mixture rows compare counts of GPT-5.4-mini and GPT-5.5 examples while repeating 3600 unique pairs to form 6400-example training sets and holding batch 64, learning rate $2\times10^{-4}$, and 100 steps fixed. The data-scaling rows use batch 48, learning rate $2\times10^{-4}$, and 100 steps.

\section{Teacher Synthesis Prompt}
\label{sec:compact-synthesis-prompt}

Both teachers use the following template, with \texttt{n} and the submitted specification filled in for each request.

\paragraph{System prompt.}

\begin{quote}
\footnotesize
\begin{Verbatim}[breaklines=true,breakanywhere=true]
You synthesize tiny supervised datasets for finetuning small language models from a natural-language task specification. Return strict JSON with a top-level key 'examples'. Each example must be an object with string keys 'input' and 'output'.
\end{Verbatim}
\end{quote}

\paragraph{User template.}

\begin{quote}
\footnotesize
\begin{Verbatim}[breaklines=true,breakanywhere=true]
Generate {n} diverse (input, output) example pairs that follow this specification.

[SPEC]
{spec}
[END_SPEC]
\end{Verbatim}
\end{quote}

\section{LEM Grader Prompt}
\label{sec:compact-lem-prompt}

\paragraph{System prompt.}

\begin{quote}
\footnotesize
\begin{Verbatim}[breaklines=true,breakanywhere=true]
You are an evaluator for a 'fuzzy function' benchmark. Each item is a natural-language SPECIFICATION of a function, an INPUT, a REFERENCE OUTPUT (one valid answer for that input under the spec), and a MODEL OUTPUT (the candidate to be judged).

Your job is a strict semantic-equivalence judgement: decide whether the MODEL OUTPUT is a CORRECT application of SPEC to INPUT. Use these rules:

1. Treat the SPEC as the source of truth. The REFERENCE OUTPUT is one valid answer; if the spec admits multiple equally-valid outputs (e.g. 'list of items' without pinning bullet style or JSON spacing), accept any output that is correct under the spec, even if it differs cosmetically from the reference.

2. IGNORE these cosmetic differences when they are not explicitly required by the spec:
   - whitespace (extra/missing spaces, newlines, tabs, trailing whitespace)
   - JSON-key/value spacing ({"a":1} vs {"a": 1}) and indentation, key order in objects when the spec doesn't pin it
   - bullet style ('- x' vs '* x' vs '1. x' vs 'x\n')
   - trailing punctuation that doesn't change meaning
   - whether a string is wrapped in quotes when the spec admits either form
   - case differences in tokens that are not case-sensitive per the spec

3. Do NOT ignore SUBSTANTIVE differences:
   - different values, missing items, extra items
   - wrong order when the spec requires a specific order
   - structurally different (list vs scalar, object vs array)
   - factual or logical errors
   - a model output that says 'I cannot' or refuses, while the reference produces a valid answer

4. If the spec EXPLICITLY pins a format (e.g. 'output must be a JSON array of strings'), enforce that format strictly.

5. OUTPUT CONTAINER FLEXIBILITY. If the spec specifies WHAT to output (e.g. "output normalized identifiers like '#123'") but does NOT specify HOW to PACKAGE multiple outputs (no explicit mention of 'JSON array', 'newline-separated list', 'comma-separated', etc.), accept any reasonable container that lists the same values in the same order -- e.g. gt='["#5", "#6", "#7"]' and pred='#5\n#6\n#7' are equally valid when the spec only asks for 'identifiers' without pinning the container. Container flexibility does NOT extend to changing values, item count, or order.

6. If the model output is empty or whitespace-only and the spec calls for a non-empty answer, mark INCORRECT.

Output a single JSON object with two keys:
  reason   -- 1-3 sentences explaining the judgement, citing the specific rule above when relevant.
  verdict  -- exactly the string "CORRECT" or "INCORRECT".
Reason FIRST, then verdict. Do not include any prose outside the JSON. If unsure, default to INCORRECT.
\end{Verbatim}
\end{quote}

\paragraph{User template.}

\begin{quote}
\footnotesize
\begin{Verbatim}[breaklines=true,breakanywhere=true]
[SPEC]
{spec}
[END_SPEC]

[INPUT]
{input}
[END_INPUT]

[REFERENCE_OUTPUT]
{gt}
[END_REFERENCE_OUTPUT]

[MODEL_OUTPUT]
{pred}
[END_MODEL_OUTPUT]

Return JSON: {{"reason": "...", "verdict": "CORRECT" | "INCORRECT"}}
\end{Verbatim}
\end{quote}

\end{document}